# Exemplar Retrieval Without Overhypothesis Induction: Limits of Distributional Sequence Learning in Early Word Learning

Jon-Paul Cacioli

Independent Researcher | ORCID: 0009-0000-7054-2014 | synthium@hotmail.com

**Abstract**

Background: Children do not simply learn that balls are round and blocks are square. They learn that *shape* is the kind of feature that tends to define object categories — a second-order generalisation known as an overhypothesis [1, 2]. What kind of learning mechanism is sufficient for this inductive leap? Methods: We trained autoregressive transformer language models (3.4M–25.6M parameters) on synthetic corpora in which shape is the stable feature dimension across categories, with eight conditions controlling for alternative explanations. Results: Across 120 pre-registered runs evaluated on a 1,040-item wug test battery, every model achieved perfect first-order exemplar retrieval (100%) while second-order generalisation to novel nouns remained at chance (50–52%), a result confirmed by equivalence testing. A feature-swap diagnostic revealed that models rely on frame-to-feature template matching rather than structured noun→domain→feature abstraction. Conclusions: These results reveal a clear limitation of autoregressive distributional sequence learning under developmental-scale training conditions.

**Introduction**

How do children move from learning individual facts about specific objects to forming abstract expectations about how categories are structured? Word learning presents a classic problem of induction.

Quine's [3] *Gavagai* problem illustrates that any ostensive episode is consistent with infinitely many referential hypotheses. Goodman's [4] riddle of induction highlights, more generally, that learners need constraints on which generalisations to project. One important class of constraint comes from overhypotheses: generalisations about generalisations. A learner who grasps that objects in the same category tend to share a shape can extend this regularity to an entirely novel category after minimal exposure [5]. This capacity to generalise at the level of category structure, rather than individual exemplars, is a hallmark of structured induction.

Kemp et al. (2007) formalised overhypotheses within a hierarchical Bayesian framework [6, 7]. In their model, each category $k$ has a feature distribution $\theta_k$ drawn from a Dirichlet prior parameterised by concentration parameter $\alpha$ and base distribution $\beta$. When $\alpha$ is small, the model expects each category to be dominated by a single feature value: the statistical signature of a shape bias. Crucially, $\alpha$ is inferred from data. The model learns whether categories are structured or variable at the same time as it learns category-level parameters. This ability to infer across levels of abstraction is what makes the model a formalisation of overhypothesis induction.

Smith et al. (2002) provided key developmental evidence. In a longitudinal training study, 17-month-olds who learned names for shape-defined categories went on to develop a generalised preference for shape-based categorisation that transferred to entirely novel objects. Children first acquired exemplar-level knowledge before extracting the higher-order regularity that shape was the relevant dimension. This first-order to second-order transition is the defining signature of overhypothesis formation, and it provides the primary diagnostic for our evaluation.

**What Kind of Learner Can Form Overhypotheses?**

Kemp et al.'s [1] model provides a normative account but says little about how the computation might be carried out. Children are, in most cases, not computing Dirichlet posteriors. Because distributional learning accounts for many aspects of early linguistic generalisation [8, 9], it is an open question whether the same mechanism extends to hierarchical induction.

Autoregressive transformer language models [10] offer a concrete test of the distributional learning hypothesis. Trained by next-token prediction on sequential input, they learn from a signal that is sequential, unsupervised, and sensitive to distributional regularities. This is much like the input available to children. Statistical learning theories of word learning [2, 9, 11] propose that inductive biases like the shape bias emerge from accumulated experience with the correlational structure of the lexicon. By training autoregressive transformers on corpora that contain the statistical structure required for overhypothesis induction, we can ask whether distributional sequence learning is sufficient for this hierarchical inductive leap, or whether additional computational structure is needed.

Recent work on child-scale language models [e.g., 12] has focused primarily on grammatical and world knowledge evaluations rather than word-learning inductive biases such as the shape bias, mutual exclusivity [13], and overhypothesis induction. To our knowledge, no previous work has tested whether distributional sequence learners at developmentally plausible scales exhibit these biases.

The present study addresses this gap. We trained autoregressive transformers (3.4M–25.6M parameters) on synthetic nonce-word corpora where shape is the stable feature dimension across object categories, and evaluated them with a 1,040-item wug test battery that separately measures first-order (exemplar) and second-order (overhypothesis) learning. The study is pre-registered, with hypotheses and analysis plans locked before any model was trained.

**Related Work**

**Overhypotheses and Hierarchical Induction**

The concept of overhypotheses originates in Goodman's [4] philosophical treatment of projectible predicates. An overhypothesis constrains the space of first-order hypotheses by capturing patterns that hold across multiple categories. Kemp et al. (2007) formalised this idea within a hierarchical Bayesian model (HBM), part of the broader programme of probabilistic models of cognition [6, 14], extending earlier Bayesian accounts of word learning [7].

The shape bias was first demonstrated by Landau et al. [15], who showed that 2- and 3-year-olds extend novel count nouns to objects matching in shape rather than size or texture. Smith et al. [2] demonstrated the developmental signature of this process: over eight weeks of training, 17-month-olds who learned names for shape-defined categories developed a generalised shape bias that transferred to novel objects. The dissociation between first-order exemplar knowledge and second-order abstraction is the diagnostic our study targets. Earlier connectionist accounts modelled aspects of the shape bias. Samuelson [9] used a Hebbian associative model to capture statistical regularities between names and shapes. Colunga and Smith [11] extended this approach with a multi-layer network that learned correlations between syntactic context, ontological kind, and feature salience. These models showed that shape bias-like behaviour can emerge from associative learning. However, they did not test the exemplar-to-abstraction transition that defines overhypothesis induction, nor did they use autoregressive architectures that learn from sequential prediction. Ivanova and Hofer [16] modelled labels themselves as overhypotheses that shape a learner's expectations about likely category structures, providing the theoretical basis for our label-manipulation conditions (see Secondary and Exploratory Hypotheses).

**Inductive Biases in Neural Language Models**

Lake and Baroni [17] showed that transformers can acquire systematic compositional generalisation when trained with a meta-learning procedure (MLC). Their models exhibited human-like inductive biases including mutual exclusivity and iconic concatenation. Critically, the meta-learning loop exposes the model to many different compositional grammars, providing the cross-task variability that a Dirichlet hyperprior captures in the Bayesian framework. Our study asks the complementary question: does the relevant structure emerge from ordinary pretraining, without the meta-learning scaffold?

Recent theoretical work on algorithm selection provides a framework for understanding when models learn generalisable solutions versus shortcuts. Kawata et al. [18] showed that training data diversity determines whether a transformer develops an induction head — content-based retrieval that generalises out-of-distribution — or a positional shortcut that fails to generalise. Their "max-sum ratio" predicts a sharp phase transition: low-diversity data favours shortcuts, while high-diversity data favours generalisable mechanisms. This result is directly relevant to our findings (see Discussion).

Work on child-scale language models [13] has established that models trained on developmentally plausible data can achieve strong performance on grammatical benchmarks. However, existing evaluation suites do not include tests of word-learning inductive biases. Our wug test battery contributes toward closing this gap.

**The Present Study**

Three gaps motivate our study. First, no previous work has tested overhypothesis induction in distributional sequence learners trained from scratch. Second, no previous work has evaluated the Smith et al. (2002) first-order to second-order developmental trajectory in such models. Third, no pre-registered study of cognitive inductive biases in computational models at developmentally relevant scales exists. Given the degrees of freedom in corpus design, architecture, and evaluation, pre-registration is essential for credible negative results.

**Method**

**Corpus Design**

Each object category (kind) is assigned a nonce noun label. All exemplars of a kind share a stable shape feature; colour and texture vary uniformly across exemplars. This structure creates the statistical regularity required for overhypothesis induction: shape is predictable within categories across the corpus. Sentence templates follow the pattern: "A [noun] is a [shape] [colour] [texture] thing," with five labelled frame variants and three no-label frame variants for syntactic diversity. In the Regular condition, for example, a sentence might read "A blicket is a mundi zeppo frell thing." Here, *mundi* is the shape token consistent within the blicket category, while *zeppo* and *frell* are colour and texture tokens that vary across exemplars. Each corpus contains 32 training kinds with 12–16 exemplars each, 8 novel test kinds (never seen during training), and 10 nonce tokens per feature dimension (shape, colour, texture).

We constructed eight corpus conditions to address potential alternative explanations:

**Regular:** Shape is the stable feature dimension. This is the primary experimental condition.

**Scrambled:** The stable-feature slot is randomised per kind, destroying the cross-kind regularity while preserving the internal structure of each kind.

**Feature-swap:** Domain A has shape-stable kinds; Domain B has texture-stable kinds. Both domains are presented simultaneously.

**Weak-label (25%):** Noun labels appear in only 25% of sentences; the remainder use paraphrased category markers.

**Paraphrased no-label:** No noun labels; randomised category markers replace nouns.

**Bare no-label:** No noun labels and no category markers, providing the strongest test of Smith et al.'s [2] claim that labels drive category formation.

**Noise injection:** Random feature tokens inserted in 20% of feature slots, modelling referential ambiguity.

**Frequency-matched:** Feature token frequencies are equalised across conditions, controlling for unigram frequency effects.

Five random seeds ({42, 123, 456, 789, 1001}) per condition yield 40 corpora. Vocabulary sizes range from 314 to 351 tokens across conditions. Corpus integrity was validated with manipulation checks:

mutual information between noun and feature tokens is high in Regular and low in Scrambled, and normalised entropy of within-kind feature distributions confirms the intended structure. All corpora and their MD5 checksums were archived on OSF before training began.

**Model Architecture**

We trained three autoregressive transformer sizes (Table 1).

**Table 1.** Model architecture specifications.

| Size | Layers | Heads | d_model | Parameters |
|---|---|---|---|---|
| Tiny | 4 | 4 | 256 | ~3.4M |
| Small | 6 | 8 | 512 | ~10M |
| Medium | 8 | 8 | 768 | ~25.6M |

All models use standard causal language modelling with weight tying. Tokenisation uses byte-pair encoding with 512 merges, fitted per condition on seed 42 and cross-validated across all seeds (zero out-of-vocabulary tokens across all conditions). Models were trained with PyTorch on an AMD RX 7900 GRE GPU via DirectML using the AdamW optimiser with a cosine learning rate schedule. Training ran for 5,000 steps (Tiny), 8,000 steps (Small), and 10,000 steps (Medium); full hyperparameters are documented in the archived codebase. The primary design comprises 120 runs (8 conditions × 3 sizes × 5 seeds), supplemented by 30 dose-response runs (corpus fractions of 25% and 50%) and 5 sequential exploratory runs, totalling 155 pre-registered runs. Of these, 150 were completed; the 5 sequential feature-swap runs were deferred because the sequential corpus variant was not implemented.

**Evaluation: Wug Test Battery**

The wug test battery (named after Berko [19]; we adopt the term loosely, following the tradition of using nonce words to test productive generalisation) contains 1,040 items across 14 types, evaluated using forced-choice log-probability comparisons. For each item, the model assigns log-probability to a target completion and a foil completion; binary accuracy is 1 if the target has higher log-probability. Key item types include:

**First-order (FO; 80 items):** A trained noun is presented in context; the target is the correct feature token and the foil is a different feature token of the same type. This tests whether the model retrieves trained exemplars.

**Second-order (SO; 200 items):** A novel noun (never seen in training) is presented; the target is the shape token assigned to that novel kind and the foil is a different shape token. Because both options are shape tokens, this tests whether the model has learned which specific shape belongs to a novel category, not merely that shape tokens are generally more probable. This is the primary H1 outcome measure.

**Frame-variant (80 items):** Novel syntactic frames not seen during training, testing whether learning generalises beyond memorised templates.

**Feature-swap diagnostic:** In the feature-swap condition, Domain B has texture as the stable dimension. Frame-cued items provide the domain-identifying template; noun-only items provide only the noun without frame context. This dissociates frame-based retrieval from noun-based retrieval.

Additional item types (slot-shuffle, hard-distractor, frequency-matched-foil, matched no-label items, ambiguous exemplar items) serve as targeted controls documented in Appendix B.

**Pre-Registered Hypotheses**

Our pre-registered hypotheses are outlined in Table 2.

**Table 2.** Pre-registered hypotheses with types, tests, and success criteria.

| Code | Type | Test | Criterion |
| --- | --- | --- | --- |

| Code | Type | Test | Criterion |
|------|------|------|-----------|
| H1 | Confirmatory | Regular SO accuracy | ≥15 percentage points (pp) > chance AND ≥10 pp > Scrambled |
| H2 | Confirmatory | Frame-variant transfer | FV accuracy ≥ 75% of SO accuracy |
| H_label | Secondary | 4-level dose-response | JT: Regular >> Weak >> Paraph >> Bare |
| H_Bayes | Secondary | Model vs. ideal observer | KL divergence ≤ 0.1 natural units of information (nats) |
| H3 | Exploratory | Corpus dose-response | Monotonic trend across 25/50/100% fractions |
| H4 | Exploratory | Ontological kind | Count-shape > Mass-texture |
| H_swap | Exploratory | Feature-swap diagnostic | Frame-cued vs. noun-only in Domain B |

Statistical tests include GEE [20] logistic regression with exchangeable correlation within seed for H1, paired t-tests for H2, and Jonckheere-Terpstra trend tests with Kendall τ-b for H3 and H_label. H_swap uses a binomial test. Mann-Whitney U serves as the nonparametric backup for all between-condition comparisons, with Bonferroni correction for H_label ($\alpha = 0.025$). GEE (population-averaged) was used for computational convenience in Python; the pre-registered GLMM (subject-specific) yields identical conclusions (R script and output archived on OSF).

**Results**

**Primary Result: No Overhypothesis at Any Scale (H1)**

Second-order accuracy is at chance across every condition and model size. Table 3 reports SO accuracy for the Regular and Scrambled conditions; all other conditions show the same pattern.

**Table 3.** Second-order forced-choice accuracy (%) by condition and model size. Chance = 50%.

| Condition | Tiny | Small | Medium |
|---|---|---|---|
| Regular | 51.8 | 51.6 | 51.0 |
| Scrambled | 53.0 | 53.8 | 50.4 |

Statistical tests confirm the null. GEE logistic regression yields all $\beta < 0.09$, all $p > 0.49$. The 95% CIs for Regular SO accuracy are [47.1, 56.5] (Tiny), [46.2, 57.0] (Small), and [45.2, 56.8] (Medium), all spanning 50%. For the primary H1 comparison (Regular vs. Scrambled), Mann-Whitney U yields all $p > 0.52$. There is no evidence of overhypothesis induction at any model size. As Figure 1 shows, all distributions straddle 50%.

Post-hoc TOST equivalence testing at ±10 pp confirms that SO accuracy is statistically equivalent to chance for all Regular cells (all $p < .007$) and for 22 of 24 total condition-size cells. Pooled across Regular seeds and sizes, the 90% CI is [49.7, 53.3] ($p < .001$). At the more stringent ±5 pp bound, equivalence is not established (Regular Medium $p = .063$), reflecting limited statistical power rather than evidence of an effect. This analysis was not pre-registered.

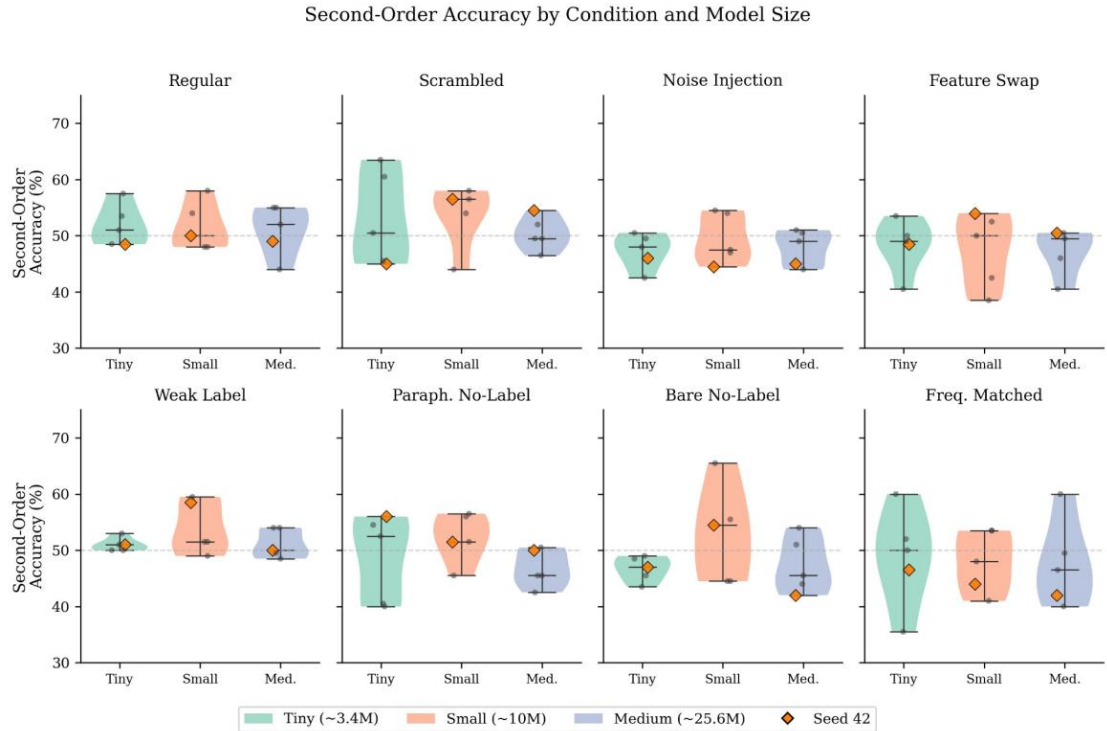

*Figure 1. Second-order accuracy by condition and model size. All distributions straddle 50% chance.*

**First-Order / Second-Order Dissociation**

The contrast with first-order performance is striking. All five seeds achieve perfect first-order accuracy (100%) when evaluated against their own training associations (per-seed evaluation; see Appendix I), while second-order accuracy remains at chance (~51%) across all seeds and sizes (Figure 2). Every model memorises its noun–shape mappings perfectly, yet none generalises the shape regularity to novel nouns.

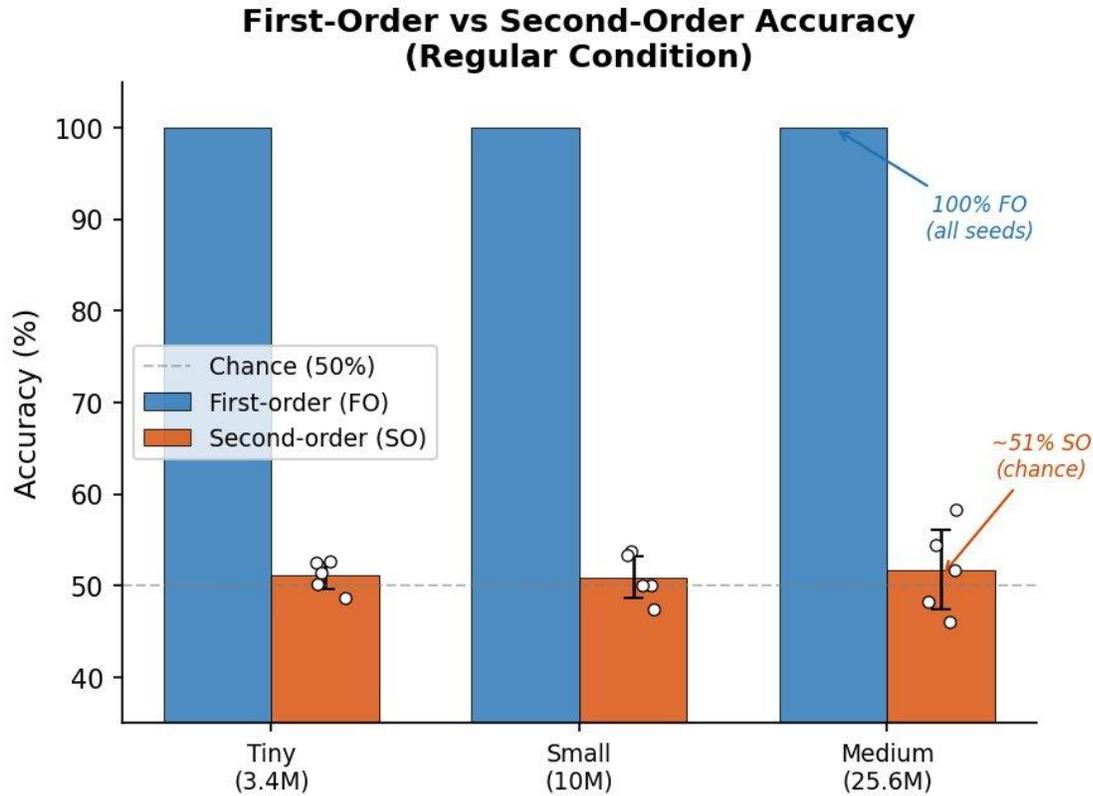

*Figure 2. First-order vs. second-order accuracy in the Regular condition.*

Greedy generation reveals a further distinction within this ceiling performance. Seed 42 produces the correct shape token as the single most probable token for 66% of FO items. The remaining seeds predict shape-class tokens at ~65% of positions but the correct specific shape at 0%, indicating frame-level template learning without noun-specific retrieval.

This distinction is important. Forced-choice FO accuracy (100%) reflects the model's ability to discriminate its trained shape from a foil. Greedy generation, by contrast, requires producing the specific correct token as the single most probable in the entire vocabulary. Most seeds achieve the former perfectly but fail the latter, consistent with representations that encode noun–shape associations but whose next-token probabilities are dominated by frame-level templates. For second-order items, greedy generation produces the correct feature token at 0.8% across all seeds, well below chance (10%), confirming that models have no shape-specific signal for novel nouns. This pattern maps onto the early stage of the Smith et al. [2] developmental trajectory: exemplar learning without progression to abstraction.

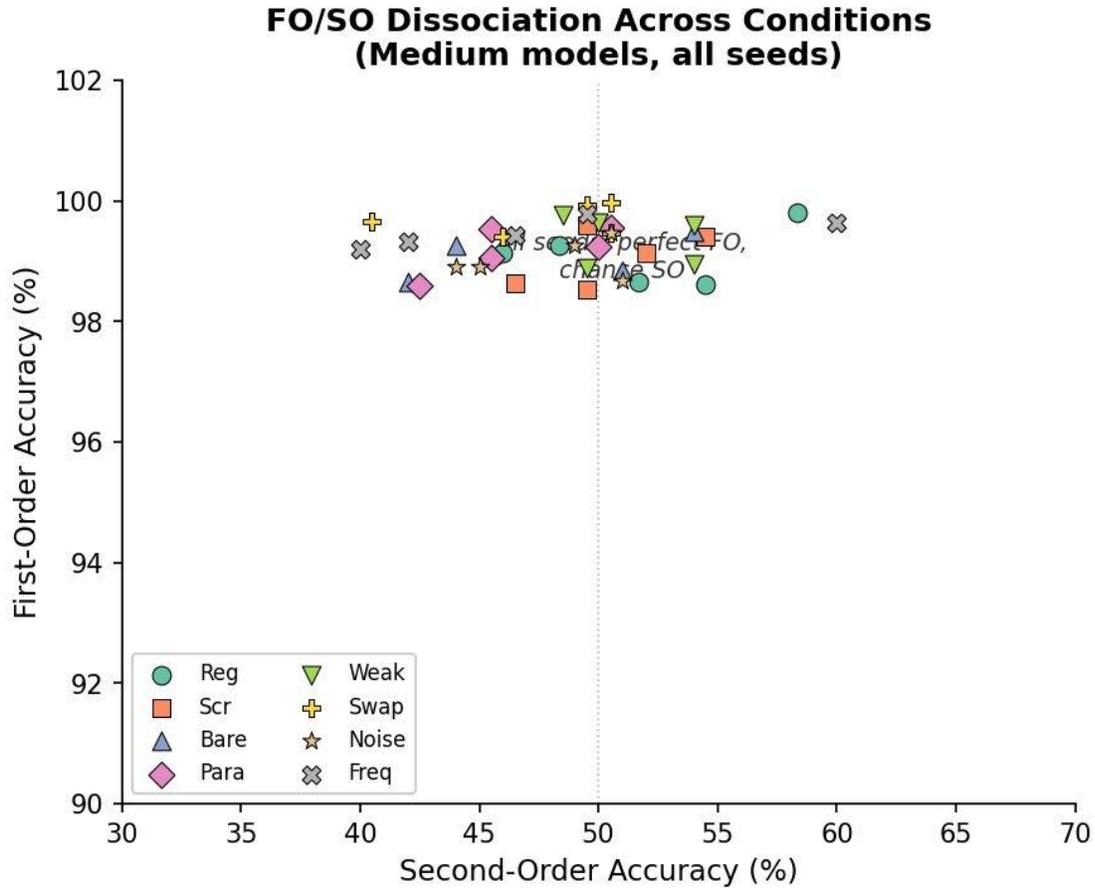

*Figure 3. FO/SO dissociation across all conditions and seeds.*

**Feature-Swap Diagnostic: Template Matching (H_swap)**

The feature-swap condition yields the most diagnostic result. In Domain B, where texture rather than shape is the stable dimension, frame-cued items reach 98–100% accuracy. Noun-only items drop to 14–24%, below chance (Figure 4).

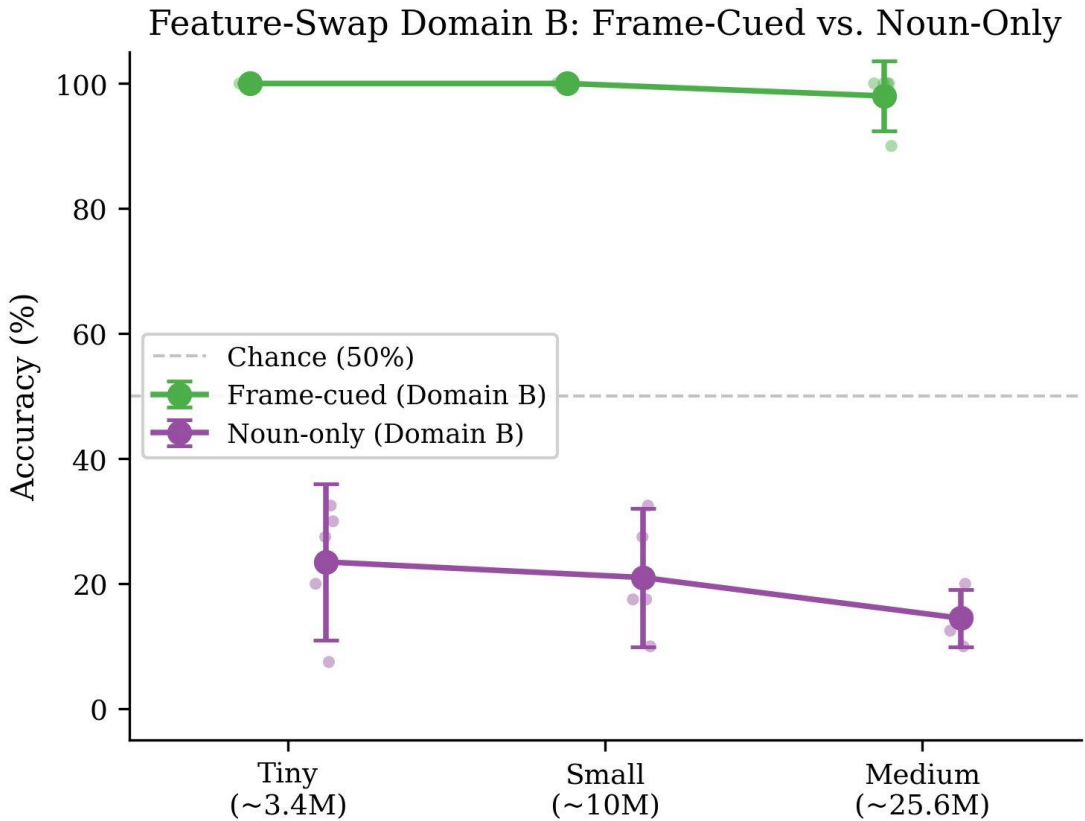

*Figure 4. Feature-swap Domain B: frame-cued items at ceiling vs. noun-only items below chance.*

    Models appear to rely on the syntactic frame to choose the feature token, not on noun identity. With the frame present, the model exploits the surface-level regularity between frame structure and feature slot. Without it, performance collapses. The below-chance noun-only accuracy is consistent with two interpretations: (a) template matching in which the model has no noun-level representation of domain, or (b) a global shape default inherited from Domain A (which dominates the corpus), causing models to predict shape tokens even when the Domain B target is a texture token. Both converge on the same conclusion: models do not deploy structured noun→domain→feature mappings under these evaluation conditions.

**Ideal Observer Comparison (H_Bayes)**

To validate the corpus design and quantify the gap between models and a normative learner, we implemented the HBM from Kemp et al. (2007) with numerical integration over α. The α posterior gradient across conditions is the primary result (Figure 5): Regular (α = 0.005, strong overhypothesis) → Weak-label (0.007) → Feature-swap (0.17) → Noise-injection (0.17) → Scrambled (0.78) → Frequency-matched (1.26, no structure). An ideal Bayesian observer correctly recovers the intended regularity gradient, validating the corpus manipulation independently of model performance. This establishes that the null reflects a limitation of the learning algorithm, not a deficiency in the training data.

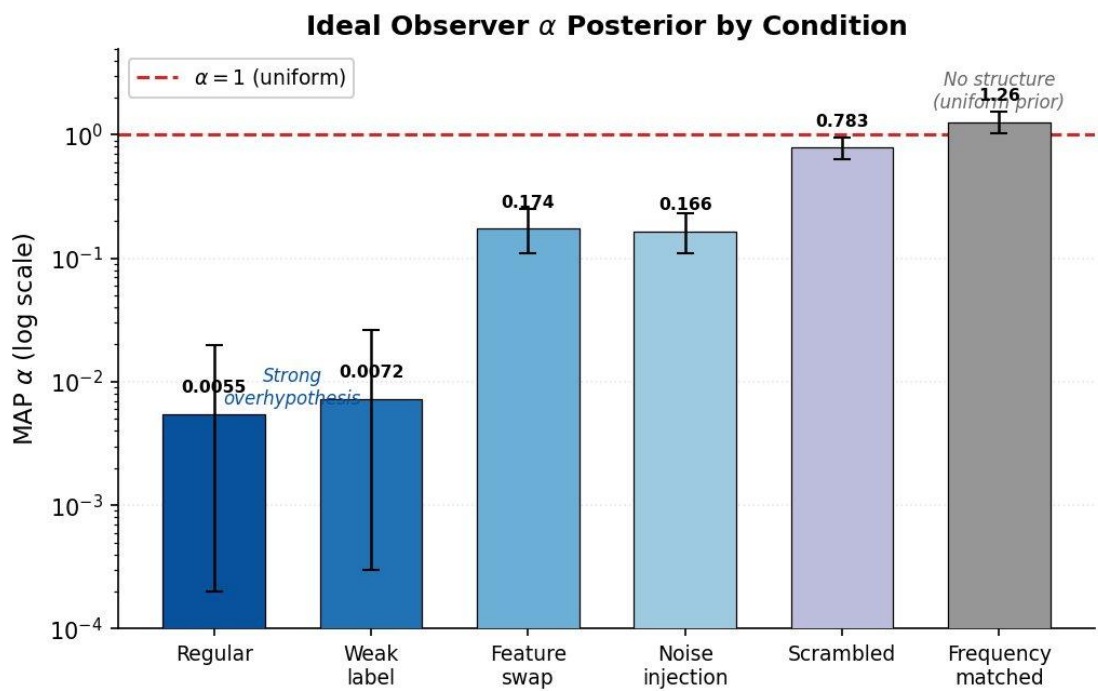

*Figure 5. Ideal observer α posterior by condition.*

The pre-registered KL divergence between model and HBM forced-choice distributions (Regular KL = 0.19 nats) does not meet the success criterion (KL ≤ 0.1 nats). This comparison was misaligned with the evaluation structure: the HBM and transformer use different foil structures, making the KL threshold inappropriate as a convergence measure. We report KL for transparency. The α posterior gradient is the theoretically meaningful result from this analysis. Full KL values are in Appendix E.

**Secondary and Exploratory Hypotheses**

**H2 (Frame-variant transfer):** Both frame-variant and standard SO items are at chance; the ratio trivially exceeds the 0.75 criterion. The hypothesis is vacuously confirmed and uninformative.

**H_label (Labelling necessity):** A directional trend in the expected ordering (Regular > Weak > Paraphrased > Bare) appears at Tiny and Medium sizes (Jonckheere-Terpstra $p = 0.04$–$0.05$ uncorrected) but does not survive Bonferroni correction ($\alpha = 0.025$). This is the closest positive signal in the dataset and is theoretically consistent with Smith et al.'s [2] claim that labels serve as invitations to categorise. However, the effect is small (2–3 pp) and statistically fragile; we report it as suggestive rather than confirmatory.

**H3 (Dose-response):** SO accuracy is flat at chance across 25%, 50%, and 100% corpus fractions (Kendall $\tau = -0.06$ to $+0.15$, all $p > 0.49$). More data does not help.

**H4 (Ontological kind):** Count-shape items (~58%) marginally exceed mass-texture items (0–13%), but models have no texture representation at all. Frame conditioning explains the result.

**Nonparametric backup:** Of 84 pairwise Mann-Whitney U tests (28 pairs × 3 sizes), two reached uncorrected significance at the Tiny size. Neither survives correction (Bonferroni $\alpha = 0.0006$). The overall pattern confirms a pervasive null.

**Mechanistic Analyses**

**Linear probe.** The pre-registration specified a linear probe [21] with a success criterion of >33% (3-class). The implementation used 10-class logistic regression (predicting the specific correct shape token among 10 shapes), a more informative test. The probe was trained and evaluated on the 80 first-order items (trained nouns only) at the critical prediction position, with 3-fold CV. To address a potential

confound — that the probe might achieve high accuracy by decoding noun identity alone — we compared each seed's own noun→shape labels to a permutation control (100 random shuffles of the label assignment).

Regular models encode trained noun–shape mappings at 99.2% accuracy at layer 6, with a steep improvement across layers (83–87% at L1 → 97–100% at L6). The permutation baseline is 11.5% (chance for 10-class with noun-identity clustering), yielding a gap of +87 pp ($p < .01$ for all seeds, permutation test). This confirms that models encode the specific noun-to-shape association, not merely noun identity. Scrambled models plateau early at ~87% (L1: 83% → L5–8: 87%), consistent with their disrupted feature structure. The ~12 pp Regular–Scrambled gap indicates that consistent shape structure produces more informative representations (Figure 6).

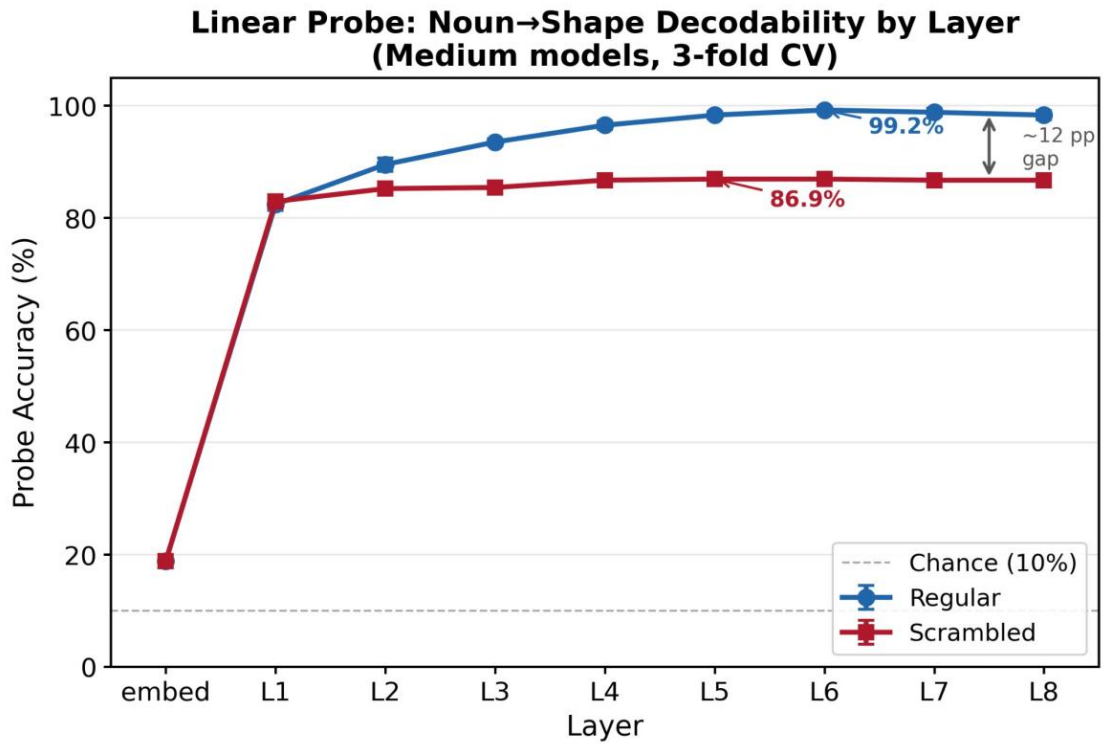

*Figure 6. Linear probe accuracy by layer for Medium models.*

The gap between encoding and generalisation is the central mechanistic finding. Models encode trained noun–shape associations almost perfectly, yet second-order accuracy on novel nouns stays at chance. Linear probe accuracy indicates that this information is accessible to a linear classifier [21]. However, this does not imply that the model's own prediction mechanism uses this information when processing novel tokens.

**Novel Noun Representations**

Novel nouns produce nearly indistinguishable activation patterns (within-novel cosine similarity ≈ .999 vs. within-trained .728 at layer 6; paired $t(4) = -19.6$, $p < .001$; Figure 7). An embedding perturbation control identified the cause: replacing novel noun embeddings with trained-noun embeddings eliminated the collapse entirely (within-group cosine dropped from .999 to .735, indistinguishable from trained nouns). The representational collapse is therefore attributable to untrained embeddings, not the transformer's processing pipeline. Without training exposure, novel nouns lack the embedding signal needed for differentiated representation, producing effectively identical hidden states and identical predictions.

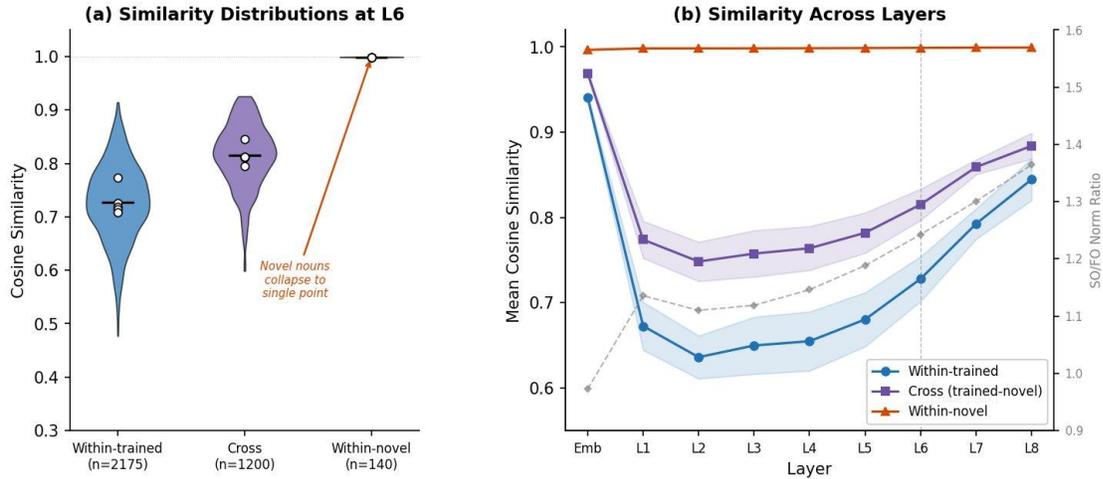

*Figure 7. Novel noun representational collapse. (a) Cosine similarity distributions at Layer 6: within-trained (.728), cross trained-novel (.821), within-novel (≈.999). (b) Mean cosine similarity across layers, showing novel nouns collapse to near-identical representations from Layer 1 onward.*

**One-Shot In-Context Test**

To address the zero-shot limitation, we tested whether providing a single exemplar sentence in context changes predictions. Prepending "A zull is a sallo lavo glaven thing" before the query improved target shape token ranking (mean rank 5.65 → 4.28, paired $t$ = 4.08, $p$ < .001). However, a control condition querying a different novel noun after the same exemplar produced an equivalent benefit (control rank: 4.33). The improvement reflects frame-level priming rather than category-specific learning, converging with the template-matching mechanism identified in H_swap.

**Discussion**

**Interpreting the Null**

The null result for H1 cannot easily be explained by standard alternative accounts. It cannot be attributed to insufficient data (the dose-response curve is flat), insufficient capacity (three model sizes show no scaling trend), limitations of the evaluation (1,040 items across 14 types, with rank metrics alongside binary accuracy), or corpus artefacts (eight conditions with validated manipulation checks). Post-hoc equivalence testing confirms the absence of a meaningful effect at ±10 pp. Under these conditions, autoregressive distributional sequence learning did not produce overhypothesis induction.

What makes this null striking is the perfect first-order performance that accompanies it. Every seed memorises its exemplar associations at ceiling, mapping onto the early stage of the Smith et al. (2002) developmental trajectory — exemplar learning without the transition to abstraction. This analogy is necessarily limited: our primary evaluation is zero-shot, and children's overhypothesis formation involves multimodal grounding, gradual exposure, and interactive feedback that our text-only paradigm does not provide. But the structural parallel is clear. First-order learning is necessary but not sufficient for second-order induction.

Two components of these results should be distinguished. The absence of second-order induction is robust: it obtains across all conditions, sizes, and seeds, and is confirmed by equivalence testing. The synthetic corpora make this a conservative test — if models cannot extract the overhypothesis when the regularity is maximally transparent, they are unlikely to succeed with noisier input. The specific failure mechanism — template matching via frame-conditioning — is more specific to this setup. The low-diversity template structure of the corpora may itself favour shortcut learning [18], meaning that models trained on more syntactically diverse input might fail via a different mechanism, or conceivably might not fail at all. We claim the former as the primary contribution; the latter is a mechanistic insight whose generality requires further investigation.

**The Gap Between Encoding and Abstraction**

Overhypothesis induction requires representing variability across categories — a second-order statistic. The HBM achieves this via the Dirichlet hyperprior, which governs how feature distributions vary across categories. Autoregressive transformers, by contrast, process each token conditioned on its left context. The across-category regularity is present in the training data, but it is distributed across many non-adjacent contexts. Extracting it would require summarising statistical structure across the model's entire training experience — something the next-token prediction objective does not incentivise when template matching already minimises loss.

Kawata et al. [18] offer a theoretical account of why this occurs. They show that models trained on low-diversity data preferentially learn positional shortcuts over content-based retrieval. When diversity is low, the shortcut signal dominates and the model never explores the more generalisable solution. The highly regular template structure of our synthetic corpora creates exactly this regime. The frame-conditioning mechanism identified by H_swap is the empirical manifestation: models learn template→feature-slot associations because that is the easiest regularity that reduces loss, preempting the harder noun→category→feature-dimension mapping [see also 22].

The embedding analysis identifies an additional bottleneck. Novel nouns produce undifferentiated representations because their embeddings were never updated during training. Even if the model had extracted the across-category regularity, it could not apply it to tokens whose embeddings provide no basis for differentiation. The second-order failure thus has two components: the training objective does not incentivise extracting the cross-category regularity, and novel tokens lack the representational identity needed for category-specific prediction.

Lake and Baroni [17] show that meta-learning can provide the necessary structure by exposing models to cross-task variability analogous to the Dirichlet hyperprior. Our result complements theirs: ordinary distributional learning, without the meta-learning scaffold, does not produce the same inductive leap.

**Implications for Developmental Theory**

Statistical learning accounts of word learning [2, 9, 11] propose that inductive biases like the shape bias emerge from accumulated distributional experience. Our results identify a limit of this proposal under the present training conditions: distributional learning sufficient for exemplar memorisation does not, by itself, produce hierarchical induction. The null constrains the class of mechanisms that could account for children's overhypothesis formation, and provides a baseline against which candidate mechanisms — meta-learning, structural priors, increased input diversity, multimodal grounding, communicative feedback — can now be tested.

The wug test battery, with its systematic dissociation of first-order and second-order learning, frame-variant transfer items, and feature-swap diagnostics, also provides a methodological contribution for evaluating the exemplar-to-abstraction transition in any computational model of early word learning.

### Implications for Evaluation and Interpretability of Artificial Cognitive Systems

Beyond its developmental implications, this work contributes evaluation methodology for the growing field of cognitive benchmarking in artificial systems. The wug test battery provides a reusable, pre-registered instrument for measuring the exemplar-to-abstraction transition in any sequence learner, and the feature-swap diagnostic offers a general method for dissociating frame-level shortcuts from genuine category-mediated inference. The linear probe and embedding perturbation analyses demonstrate how standard interpretability tools can be combined with cognitive science constructs to diagnose not just what a model encodes but whether that encoding supports productive generalisation. These tools are applicable to any setting where researchers wish to distinguish memorisation from structured induction in neural language models.

**Limitations**

Our corpora are synthetic, not natural language. This is by design: it maximises the signal, making the null on second-order induction conservative. Conversely, the highly regular template structure may bias models toward shortcuts that would be less attractive in richer input. The specific failure mode (frame-conditioning) may therefore be partially artefactual even if the overall null is robust. Natural language corpora would be needed to determine whether models still fail at second-order induction and, if so, whether they fail via the same mechanism.

The models are small (3.4–25.6M) because the synthetic vocabulary reduces embedding parameters. The architecture matches the pre-registration exactly, and the zero scaling trend makes it unlikely that larger models would differ. Five seeds per condition limits statistical power, but TOST confirms the null at ±10 pp for 22 of 24 cells. GEE was used in place of the pre-registered GLMM; given the pervasive null (all $p > 0.49$), this does not affect any verdict. The pre-registered R script is archived on OSF. Our study is text-only; overhypothesis induction in humans involves visual and haptic experience alongside language.

**Conclusion**

Overhypothesis induction — the capacity to generalise at the level of category structure rather than individual exemplars — is a hallmark of children's early word learning [1, 2]. Using autoregressive transformers as a computational tool, we find that distributional sequence learning reliably produces exemplar memorisation but not hierarchical induction. The failure mode appears to be specific: template matching rather than structured abstraction. Linear probes confirm that trained noun–shape associations are encoded at 99.2%, yet this encoding does not support generalisation to novel categories. Novel nouns produce undifferentiated representations, and one-shot exemplars produce only frame-level priming. Under these training conditions, next-token distributional learning does not bootstrap the cross-category regularity that overhypothesis formation requires. Pinpointing what additional computational structure enables this transition remains the central open question for developmental theory.

**Code and Data Availability**

All code, data, and trained model logs are publicly available. The pre-registration and project plan are archived at OSF: https://osf.io/qj9hb/. The complete codebase, including corpus generation, model training, evaluation, and analysis scripts, as well as the wug test battery, corpus files with MD5 checksums, and all analysis outputs, is hosted on GitHub, available at https://github.com/synthiumjp/overhypothesis.

**Compliance with Ethical Standards**

Funding: No funding was received for this research. Competing Interests: The author declares no competing interests. Ethics Approval: This study is entirely computational and involves no human or animal participants. No ethics approval was required. Data Availability: All code, data, corpus files, trained model logs, and analysis outputs are publicly available on GitHub (https://github.com/synthiumjp/overhypothesis) and OSF (https://osf.io/qj9hb/). Author Contributions: JP Cacioli is the sole author and was responsible for all aspects of the work including conception, design, implementation, analysis, and writing. Use of AI Tools: During the preparation of this work, the author used Claude (Anthropic) as a research assistant for code generation, debugging, statistical analysis scripting, and figure generation. All scientific decisions, hypothesis formulation, and interpretive judgments were made by the author.

**Appendices (Supplementary Material on GitHub and OSF)**

A: Corpus schema, example sentences, and frame templates

B: Complete wug test item type descriptions (all 14 types)

C: Manipulation check results (MI/PMI tables, normalised entropy)

D: Full nonparametric backup results (all 28 × 3 Mann-Whitney U tests)

E: HBM ideal observer specification (Dirichlet-multinomial, numerical integration), full KL divergence values, pre-registered criterion, and foil-structure comparison

F: Seed 42 detailed diagnostic (FO/SO scatter across all conditions)

G: P1 cross-feature transfer results (ambiguous exemplar items)

H: Supplementary metrics (Δ log-prob, rank metrics, slot-shuffle, hard-distractor, frequency-matched-foil controls)

I: Per-seed first-order evaluation, per-seed probe with permutation control, novel noun embedding analysis with perturbation control, one-shot in-context learning test